\documentclass[twocolumn]{article}
\usepackage[accsupp]{axessibility}
\usepackage[pagenumbers]{cvpr} % To force page numbers, e.g. for an arXiv version

\usepackage{algorithm}
\usepackage{algorithmic}

\usepackage{comment}
\usepackage{amssymb} % define this before the line numbering.

\usepackage{graphicx}
\usepackage{booktabs}
\usepackage{multirow}
\usepackage{float}
\usepackage{multicol}
\usepackage{wrapfig}
\usepackage[english]{babel}
\usepackage{bm}
\usepackage[pagebackref,breaklinks,colorlinks]{hyperref}
\usepackage{amsmath}
\usepackage[switch]{lineno}
\usepackage{bbm}

\usepackage{soul}
\usepackage{color}
\usepackage{xcolor}
\usepackage{colortbl} 

\newcommand{\wh}[1]{\textcolor{black}{#1}}
\newcommand{\zjh}[1]{\textcolor{black}{#1}}

\usepackage[pagebackref,breaklinks,colorlinks]{hyperref}

\usepackage[capitalize]{cleveref}
\crefname{section}{Sec.}{Secs.}
\Crefname{section}{Section}{Sections}
\Crefname{table}{Table}{Tables}
\crefname{table}{Tab.}{Tabs.}

\def\cvprPaperID{6459} % *** Enter the CVPR Paper ID here
\def\confName{CVPR}
\def\confYear{2023}

\begin{document}

%%%%%%%%% TITLE - PLEASE UPDATE
\title{Actionlet-Dependent Contrastive Learning for Unsupervised Skeleton-Based Action Recognition}

\author{Lilang Lin, Jiahang Zhang, Jiaying Liu\thanks{Corresponding author. This work is supported by the National Natural Science Foundation of China under contract No.62172020.}\\
Wangxuan Institute of Computer Technology, Peking University, Beijing, China\\
% Institution1 address\\
% {\tt\small firstauthor@i1.org}
% For a paper whose authors are all at the same institution,
% omit the following lines up until the closing ``}''.
% Additional authors and addresses can be added with ``\and'',
% just like the second author.
% To save space, use either the email address or home page, not both
% \and
% Second Author\\
% Institution2\\
% First line of institution2 address\\
% {\tt\small secondauthor@i2.org}
}
\maketitle

%%%%%%%%% ABSTRACT
% \begin{abstract}
%    The ABSTRACT is to be in fully justified italicized text, at the top of the left-hand column, below the author and affiliation information.
%    Use the word ``Abstract'' as the title, in 12-point Times, boldface type, centered relative to the column, initially capitalized.
%    The abstract is to be in 10-point, single-spaced type.
%    Leave two blank lines after the Abstract, then begin the main text.
%    Look at previous CVPR abstracts to get a feel for style and length.
% \end{abstract}

\begin{abstract}
The self-supervised pretraining paradigm has achieved great success in skeleton-based action recognition. However, these methods treat the motion and static parts equally, and lack an adaptive design for different parts, which has a negative impact on the accuracy of action recognition. To realize the adaptive action modeling of both parts, we propose an \textbf{Act}ionlet-Dependent \textbf{C}ontrastive \textbf{L}ea\textbf{r}ning method (ActCLR). The actionlet, defined as the discriminative subset of the human skeleton, effectively decomposes motion regions for better action modeling. In detail, by contrasting with the static anchor without motion, we extract the motion region of the skeleton data, which serves as the actionlet, in an unsupervised manner. Then, centering on actionlet, a motion-adaptive data transformation method is built. Different data transformations are applied to actionlet and non-actionlet regions to introduce more diversity while maintaining their own characteristics. Meanwhile, we propose a semantic-aware feature pooling method to build feature representations among motion and static regions in a distinguished manner. Extensive experiments on NTU RGB+D and PKUMMD show that the proposed method achieves remarkable action recognition performance. More visualization and quantitative experiments demonstrate the effectiveness of our method. Our project website is available at \url{https://langlandslin.github.io/projects/ActCLR/}

\end{abstract}

\section{Introduction}
\label{sec:intro}

Skeletons represent human joints using 3D coordinate locations. 
Compared with RGB videos and depth data, skeletons are lightweight, \wh{privacy-preserving}, and compact to represent human motion. 
On account of being easier and more discriminative for analysis, skeletons have been widely used in action recognition task~\cite{zhang2020context,liu2020disentangling,zhang2020semantics,song2020stronger,peng2020learning,su2020predict}.

\begin{figure}[tb]
\includegraphics[width=\linewidth]{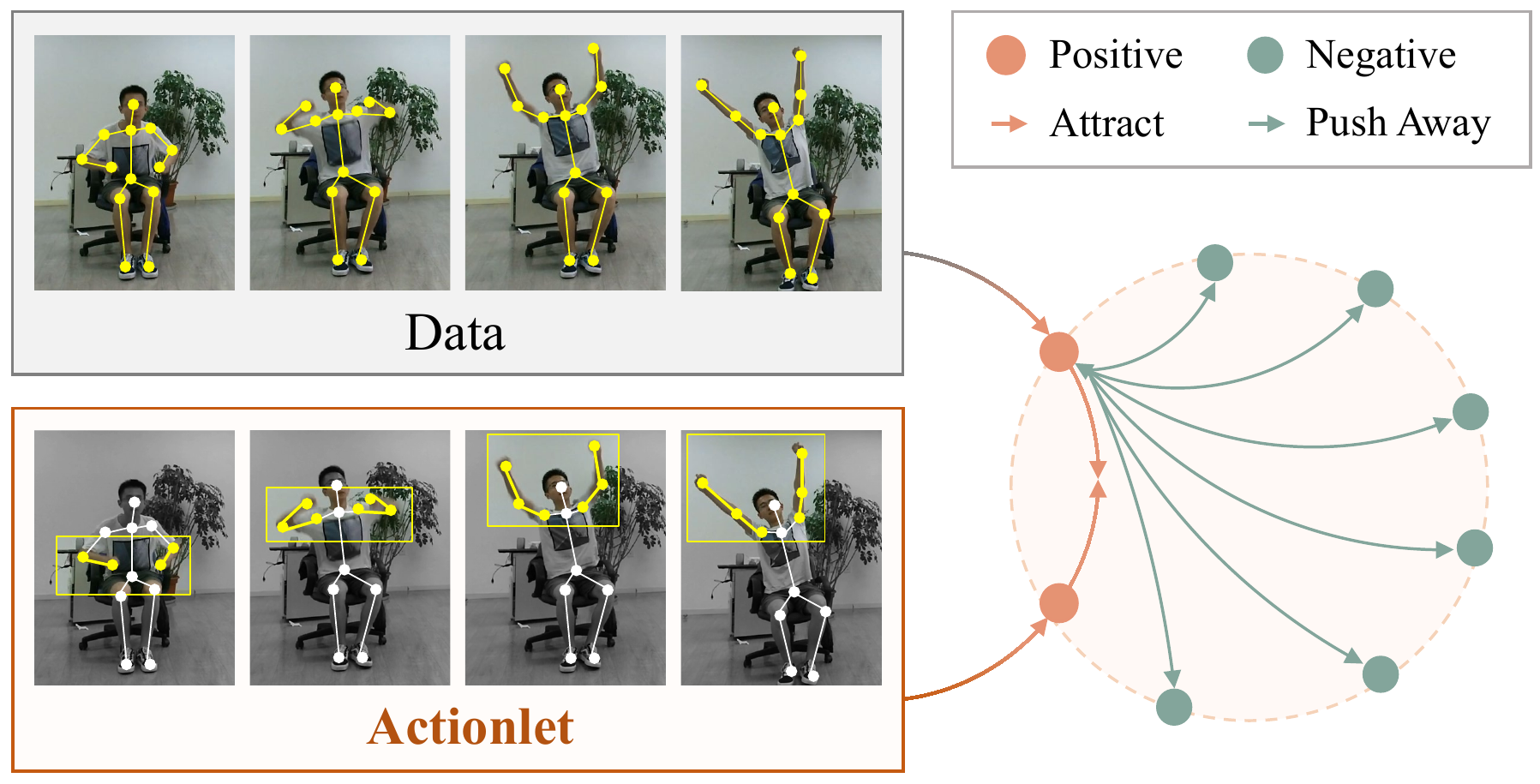}
\caption{
Our proposed approach (ActCLR) locates the motion regions as actionlet to guide contrastive learning.
}
\label{fig:teaser}
\end{figure}

Supervised skeleton-based action recognition methods~\cite{si2019attention,shi2019two,chen2021channel} \wh{have achieved impressive performance}.
However, \wh{their success highly depends} on \wh{a large amount of} labeled training data, which \wh{is} expensive to obtain. 
To get rid of the reliance on full supervision, self-supervised learning~\cite{zheng2018unsupervised,lin2020ms2l,su2020predict,thoker2021skeleton} has been introduced into skeleton-based action recognition. 
It adopts a two-stage paradigm, \wh{\textit{i.e.}} first \wh{applying} pretext tasks for unsupervised pretraining and then \wh{employing} downstream tasks for finetuning.

\wh{According to learning paradigms, all} methods can be classified into two categories: reconstruction-based~\cite{su2020predict,yang2021skeleton,kim2022global} and contrastive \wh{learning-based}.
Reconstruction-based methods \wh{capture} the \wh{spatial-temporal} correlation by predicting masked skeleton data. 
Zheng \wh{\textit{et al.}}~\cite{zheng2018unsupervised} first proposed \wh{reconstructing} masked skeletons for \wh{long-term} global motion dynamics.
 Besides, the contrastive learning-based methods have shown remarkable potential recently. These methods employ skeleton transformation to generate positive/negative samples.
Rao \wh{\textit{et al.}}~\cite{rao2021augmented} applied Shear and Crop as data augmentation.
Guo \wh{\textit{et al.}}~\cite{guo2021contrastive} further proposed to use more augmentations, \wh{\textit{i.e.}} rotation, masking\wh{,} and flippling, to improve the consistency of contrastive learning.

\wh{These} contrastive learning works treat different regions of the skeleton sequences \wh{uniformly}.
\wh{However, the motion regions contain richer action information and contribute more to action modeling.}
Therefore, it is sub-optimal to directly apply data transformations to all regions in the previous works, \wh{which may degrade the motion-correlated information too much}. 
For example, if the mask transformation is applied to the hand joints in the hand raising action, the motion information of the hand raising is \wh{totally} impaired.
It will give rise to the false positive problem, \textit{i.e.}, the semantic inconsistency due to the information loss between positive pairs.
Thus, \wh{it is necessary to adopt a distinguishable design} for motion and static regions in the data sequences.
%
% And the reconstruction task reconstructs random areas. 
% %
% But these regions may not be the region where the motion occurs. 
% %
% Therefore, it encodes a lot of motion-independent information in the model.
%
% Thus, a sequence of actions exists motion-correlated regions and static regions. It is not reasonable to treat these regions equally.

To tackle these problems, we propose a new actionlet-dependent contrastive learning method (ActCLR) by \wh{treating} motion and static regions \wh{differently}, as shown in Fig.~\ref{fig:teaser}.
An \textit{actionlet}~\cite{wang2012mining} is \wh{defined as} a conjunctive structure of skeleton joints. \wh{It is expected to be highly representative of one action and highly discriminative to distinguish the action from others.}
\wh{The actionlet in previous works is defined in a supervised way, which relies on action labels and has a gap with the self-supervised pretext tasks.}
\wh{To this end, in the unsupervised learning context, we propose to obtain actionlet by comparing the action sequence with the average motion to guide contrastive learning. In detail,} the average motion is \wh{defined as} the average of all the series in the dataset.
Therefore, this average motion is employed as the static anchor without motion. 
We contrast the action sequence with the average motion to get the area with the largest difference. 
This region is considered \wh{to be} the region where the motion \wh{takes place}, \textit{i.e.}, actionlet.

Based on this actionlet, we design \wh{a} motion-adaptive transformation strategy. 
The actionlet region is transformed by performing the proposed semantically \wh{preserving} data transformation.
\wh{Specifically,} we \wh{only} apply \wh{stronger data} transformations to non-actionlet \wh{regions}. \wh{With less interference in the motion regions, this} motion-adaptive transformation strategy makes \wh{the model learn} better semantic consistency and obtain stronger generalization performance.
\wh{Similarly,} we utilize a semantic-aware feature pooling method\wh{.} By extracting the features in the actionlet region, \wh{the features can be more representative of the motion without the interference of the semantics in static regions.}
%
% Finally, we utilize a regularization constraint based on entropy minimization for training. The entropy of the feature space is reduced during the optimization process to obtain a more \wh{definitive} output.

%
We provide thorough experiments and detailed analysis on NTU RGB+D~\cite{shahroudy2016ntu,liu2019ntu} and PKUMMD~\cite{liu2020pku} datasets to prove the superiority of our method. 
Compared to the state-of-the-art methods, our model achieves remarkable results with self-supervised learning.

In summary, our contributions \wh{are summarized as follows}:
\begin{itemize}%[leftmargin=*]
\setlength\itemsep{.3em}
\item We propose a novel unsupervised actionlet-based contrastive learning method. Unsupervised actionlets are mined as skeletal regions that are the most discriminative compared with the static anchor, \textit{i.e.}, the average motion of \wh{all} training data.
\item A motion-adaptive transformation strategy is designed for contrastive learning. In the actionlet region, we employ semantics-preserving data transformations to learn semantic consistency. And in non-actionlet regions, we apply \wh{stronger} data transformations to obtain stronger generalization performance.
\item We utilize \wh{semantic-aware} feature pooling to extract motion features of the actionlet regions. It makes features to be more focused on motion \wh{joints} without being distracted by motionless joints.
\end{itemize}

\section{Related Work}
In this section, we first introduce the related work of skeleton-based action recognition, and then briefly review contrastive learning.

\subsection{Skeleton-Based Action Recognition}
Skeleton-based action recognition is a fundamental yet challenging field in computer vision research. Previous skeleton-based motion recognition methods are usually realized with the geometric relationship of skeleton joints~\cite{vemulapalli2014human,vemulapalli2016rolling,goutsu2015motion}. The latest methods pay more attention to deep networks. Du \etal~\cite{du2015hierarchical} applied a hierarchical RNN to process body keypoints. 
% To learn the mappings between co-occurrence of joints and the human action, \wh{in} \cite{zhu2015co}, a model utilizing co-occurring joints \wh{is designed} as a strong discriminative feature and using a connection matrix. 
Attention-based methods are proposed to automatically select important skeleton joints~\cite{song2017end,zhang2018adding,song2018spatio,si2019attention} and video frames~\cite{song2017end,song2018spatio} to learn more adaptively about the simultaneous appearance of skeleton joints. However, recurrent neural networks often suffer from gradient vanishing~\cite{hochreiter2001gradient}, which may cause optimization problems. Recently, graph convolution networks attract more attention for skeleton-based action recognition. To extract both the spatial and temporal structural features from skeleton data, Yan \etal~\cite{yan2018spatial} proposed spatial-temporal graph convolution networks. To make the graphic representation more flexible, the attention mechanisms are applied in~\cite{si2019attention,shi2019two,chen2021channel} to adaptively capture discriminative features based on spatial composition and temporal dynamics.

% These supervised models have achieved excellent performance on skeleton-based action recognition. However, they rely heavily on massive data with annotated action labels. To relieve the data limitation, self-supervised methods are developed, which only require a significant number of unlabeled samples. Our work is also inspired to apply self-supervised learning for action recognition.

\subsection{Contrastive Learning}
\label{sec:contra}
Contrastive representation learning can date back to \cite{hadsell2006dimensionality}. The following approaches~\cite{tian2019contrastive,wu2018unsupervised,bachman2019learning,ye2019unsupervised,isola2015learning} learn representations by contrasting positive pairs against negative pairs to make the representations between positive pairs more similar than those between negative pairs. Researchers mainly focus on how to construct pairs to learn robust representations. SimCLR proposed by Chen \etal~\cite{chen2020simple} uses a series of data augmentation methods, such as random cropping, Gaussian blur and color distortion to generate positive samples. He \etal~\cite{he2020momentum} applied a memory module that adopts a queue to store negative samples, and the queue is constantly updated with training. In self-supervised skeleton-based action recognition, contrastive learning has also attracted the attention of numerous researchers. Rao \etal~\cite{rao2021augmented} applied MoCo for contrastive learning with a single stream. To utilize cross-stream knowledge, Li \etal~\cite{li20213d} proposed a multi-view contrastive learning method and Thoker \etal~\cite{thoker2021skeleton} employed multiple models to learn from different skeleton representations. Guo \etal~\cite{guo2021contrastive} proposed to use more extreme augmentations, which greatly improve the effect of contrastive learning. Su \etal~\cite{su2021self} proposed novel representation learning by perceiving motion consistency and continuity. 
Following MoCo v2~\cite{he2020momentum}, they exploit InfoNCE loss to optimize contrastive learning:
\begin{equation}
\label{equ:info}
\begin{aligned}
\mathcal{L}_{\text{CL}} = - \log \frac{\exp(\text{sim}(\mathbf{z}^i_q, \mathbf{z}^i_k) / \tau)}{\exp(\text{sim}(\mathbf{z}^i_q, \mathbf{z}^i_k)/ \tau) + K},
\end{aligned}
\end{equation}
where $\mathbf{z}^i_q = g_q(f_q(\mathbf{X}^i_q))$ and $\mathbf{z}^i_k = g_k(f_k(\mathbf{X}^i_k))$. $K = \sum_{j=1}^M \exp(\text{sim}(\mathbf{z}^i_q, \mathbf{m}^j)/ \tau)$ and $\tau$ is a temperature hyper-parameter. $f_q(\cdot)$ is an online encoder and $f_k(\cdot)$ is an offline encoder. $g_q(\cdot)$ is an online projector and $g_k(\cdot)$ is an offline projector. The offline encoder $f_k(\cdot)$ is updated by the momentum of the online encoder $f_q(\cdot)$ by $f_k \leftarrow \alpha f_k + (1 - \alpha) f_q$, where $\alpha$ is a momentum coefficient. $\mathbf{m}^j$ is the negative sample, stored in memory bank $\mathbf{M}$. $\text{sim}(\cdot,\cdot)$ is the cosine similarity.

\section{Actionlet-Based Unsupervised Learning}
In this section, we introduce unsupervised actionlet for contrastive representation learning, which is based on MoCo v2 described in Sec.~\ref{sec:contra}.
First, we describe the unsupervised actionlet extraction method. Then, the motion-adaptive data transformation and the semantic-aware feature pooling are introduced.

\subsection{Unsupervised Actionlet Selection}
\label{sec:actionlet}
\begin{figure*}[tb]
\begin{center}
\includegraphics[width=\textwidth]{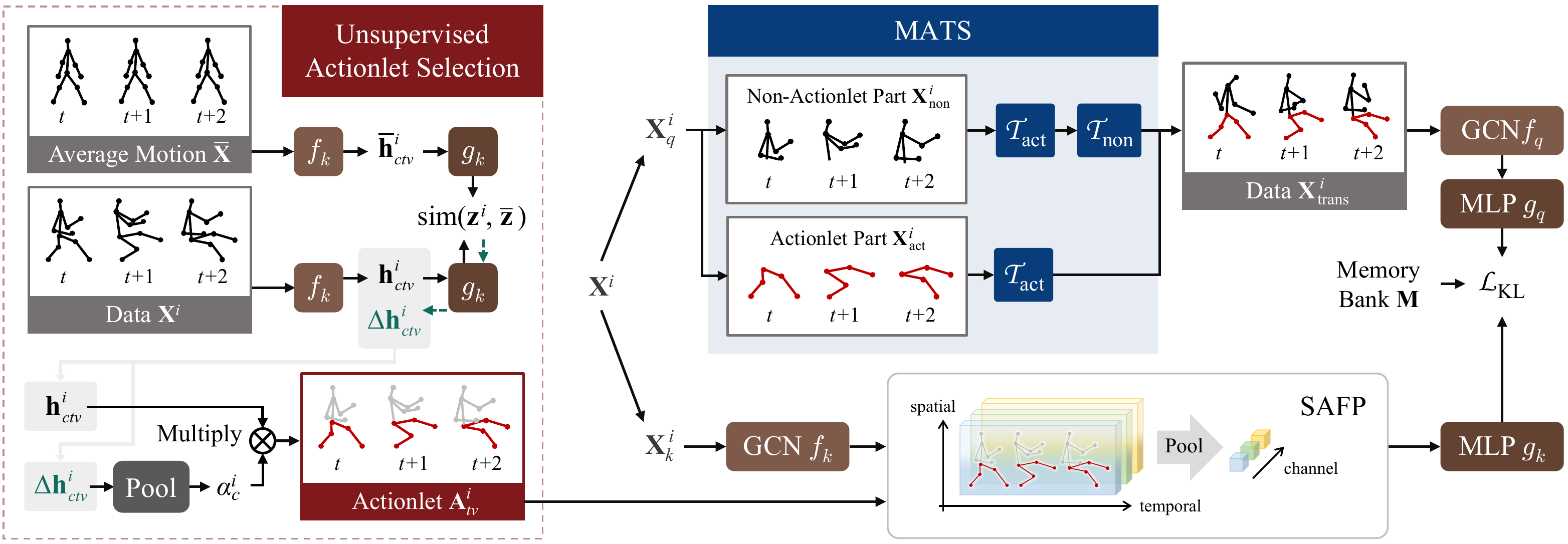}
\end{center}
\caption{
The pipeline of actionlet-dependent contrastive learning. In unsupervised actionlet selection, we employ the difference from the average motion to obtain the region of motion. For contrastive learning, we employ two streams, \textit{i.e.}, the online stream and the offline stream. The above stream is the online stream, which is updated by gradient. The below is the offline stream, which is updated by momentum. We get the augmented data $\mathbf{X}^i_{\text{trans}}$ by performing motion-adaptive data transformation (MATS) on the input data $\mathbf{X}^i_q$ with the obtained actionlet. In offline feature extraction, we employ semantic-aware feature pooling (SAFP) to obtain the accurate feature anchor. Finally, utilizing similarity mining, we increase the similarity between positives and decrease the similarity between negatives.
}
\label{fig:overview}
\end{figure*}

Traditional actionlet mining methods rely on the action label to identify the motion region, which cannot be employed in the unsupervised learning context.
Inspired by contrastive learning, we propose an unsupervised spatio-temporal actionlet selection \wh{method} to mine the motion region as shown in Fig.~\ref{fig:overview}.
The actionlet is \wh{obtained} by comparing the differences between \wh{an} action sequence and the static sequence where we assume no motion takes place.

Specifically, we introduce the average motion as the static anchor, which is regarded as the sequence without motion. Resort to this, we contrast the action sequences between the static anchor to realize actionlet localization. The details of the proposed method are described below.
\vspace{1mm}

\noindent\textbf{Average Motion as Static Anchor.}
In the process of obtaining the sequence without action occurrence, we observe that most of the action sequences have no action in most of the regions. The motion usually occurs in a small localized area, such as the hand or head.
\wh{Therefore, as shown in Fig.~\ref{fig:mean}, we can easily obtain the static anchor via average all the actions in the dataset, since most of the sequence has no motion in most of the regions and this average is a relatively static sequence.} 
It is formalized as:
\begin{equation}
    \label{equ:info}
    \begin{aligned}
    \Bar{\mathbf{X}} = \frac{1}{N}\sum_{i=1}^N (\mathbf{X}^i),
    \end{aligned}
\end{equation}
where $\mathbf{X}^i$ is the $i^{th}$ skeleton sequence and $N$ is the size of the dataset.
\vspace{1mm}
%It can also be seen in Fig.~\ref{fig:mean} that the average motion is a relatively static sequence of motions. In this case, the average motion can be viewed as a static anchor without any movement.

\noindent\textbf{Difference Activation Mapping for Actionlet Localization.}
To obtain the region where the motion \wh{takes place}, we input the skeleton sequence $\mathbf{X}^i$ with the average motion $\Bar{\mathbf{X}}$ into the offline encoder $f_k(\cdot)$ to obtain its corresponding dense features $\mathbf{h}^i_{ctv} = f_k(\mathbf{X}^i)$ and $\Bar{\mathbf{h}}_{ctv} = f_k(\Bar{\mathbf{X}})$, where $c$ means channel dimension, $t$ temporal dimension\wh{,} and $v$ joint dimension.
After global average pooling (GAP), we then apply the offline projector $g_k(\cdot)$ to obtain global features $\mathbf{z}^i = g_k(\text{GAP}(\mathbf{h}^i_{ctv}))$ and $\Bar{\mathbf{z}} = g_k(\text{GAP}(\Bar{\mathbf{h}}_{ctv}))$.
Then we calculate the cosine similarity of these two features. 
This can be formalized as:
\begin{equation}
    \label{equ:cos}
    \begin{aligned}
    \text{sim}(\mathbf{z}^i, \Bar{\mathbf{z}}) = \frac{\langle \mathbf{z}^i, \Bar{\mathbf{z}} \rangle}{\|\mathbf{z}^i\|_2\|\Bar{\mathbf{z}}\|_2},
    \end{aligned}
\end{equation}
where $\langle \cdot, \cdot \rangle$ is the inner product. 

To find the region where this similarity can be reduced, we back-propagate and reverse the gradient of this similarity to the dense feature $\mathbf{h}^i_{ctv}$. 
These gradients then are global average pooled over the temporal and joint dimensions to obtain the neuron importance weights $\alpha^i_c$:
\begin{equation}
    \label{equ:sim}
    \begin{aligned}
    \Delta \mathbf{h}^i_{ctv} & = \frac{\partial (-\text{sim}(\mathbf{z}^i, \Bar{\mathbf{z}}))}{\partial \mathbf{h}^i_{ctv}},\\
    \alpha^i_c &= \frac{1}{T \times V} \sum_{t=1}^T \sum_{v=1}^V \sigma (\Delta \mathbf{h}^i_{ctv}),
    \end{aligned}
\end{equation}
where $\sigma(\cdot)$ is the activation function.

These importance weights capture the magnitude of the effect of each channel dimension on the final difference. Therefore, these weights $\alpha^i_c$ are considered difference activation mapping. 
We perform a weighted combination of the difference activation mapping and dense features as follows:
\begin{equation}
    \label{equ:wc}
    \begin{aligned}
    \mathbf{A}^i_{tv} = \sigma\left(\sum_{c=1}^C \alpha^i_c \mathbf{h}^i_{ctv}\right) \mathbf{G}_{vv},
    \end{aligned}
\end{equation}
where $\sigma(\cdot)$ is the activation function and $\mathbf{G}_{vv}$ is the adjacency matrix of skeleton data for importance smoothing. The linear combination of maps selects features that have a negative influence on the similarity. The actionlet region is the area where the value of the
generated actionlet $\mathbf{A}^i_{tv}$ exceeds a certain threshold, while the
non-actionlet region is the remaining part.
%
% Therefore, the large value in $\mathbf{H}^i_{tv}$ corresponds to the region with the large difference from the average motion. We select the areas that exceed a certain threshold as actionlets:
% \begin{equation}
%     \label{equ:actionlet}
%     \begin{aligned}
%     \mathbf{A}^i_{tv} = \frac{\text{Clip}(\mathbf{H}^i_{tv}, [0, k])}{k},
%     \end{aligned}
% \end{equation}
% where $\text{Clip}$ is employed to resize $\mathbf{H}^i_{tv}$ into $[0,1]$ and $k$ is the threshold of activation. 

% Grad-CAM~\cite{selvaraju2017grad} also applies gradient back-propagation. In Grad-CAM, the activation regions are visualized according to the categories. 
%
% Instead, we mine the region that differs most from the average motion without any label reliance. The region that has a large difference compared to it can be considered as the region where the motion occurs, \textit{i.e.}, actionlet.

\subsection{Actionlet-Guided Contrastive Learning}

To take full advantage of the actionlet, we propose an actionlet-dependent contrastive learning method, shown in Fig.~\ref{fig:overview}.
We impose different data transformations for different regions by a motion-adaptive data transformation strategy module (MATS). 
Moreover, the semantic-aware feature pooling module (SAFP) is proposed to aggregate the features of actionlet region for \wh{better} action modeling.
\vspace{1mm}

\noindent\textbf{Motion-Adaptive Transformation Strategy (MATS).} 
\label{sec:mats}
In contrastive learning, data transformation $\mathcal{T}$ is crucial for semantic information extraction and generalization capacity. 
How to design more diverse data transformations while maintaining relevant information for downstream tasks is still a challenge.
Too simple data transformation is limited in numbers and modes and cannot obtain rich augmented patterns. However, data transformations that are too difficult may result in loss of motion information.
\zjh{To this end}, we propose motion-adaptive data transformations for skeleton data based on actionlet.
For different regions, we propose two transformations, actionlet transformation and non-actionlet transformation.

\vspace{1mm}

$\bullet$ \textbf{Actionlet Transformation $\mathcal{T}_{\text{act}}$}: 
Actionlet data transformations are performed within the actionlet regions. Inspired by the previous work~\cite{guo2021contrastive}, we adopt four spatial data transformations \{Shear, Spatial Flip, Rotate, Axis Mask\}; two temporal data transformations \{Crop, Temporal Flip\}; and two spatio-temporal data transformations \{Gaussian Noise, Gaussian Blur\}.

Besides, Skeleton AdaIN is proposed as a mixing method of global statistics. 
We randomly select two skeleton sequences and then swap the spatial mean and temporal variance of the two sequences. 
This transformation is widely used in style transfer~\cite{huang2017arbitrary}.
Here, we are inspired by the idea of style and content decomposition in style transfer and regard the motion-independent information as style and the motion-related information as content. Therefore, we use Skeleton AdaIN to transfer this motion independent noise between different data.
The noisy pattern of the data is thus augmented by this transfer method.
This transformation can be formalized as:
\begin{equation}
    \label{equ:info}
    \begin{aligned}
    \mathbf{X}^i_{\text{adain}} = \sigma(\mathbf{X}^j) \left(\frac{\mathbf{X}^i - \mu(\mathbf{X}^i)}{\sigma(\mathbf{X}^i)}\right) + \mu(\mathbf{X}^j),
    \end{aligned}
\end{equation}
where $\sigma(\cdot)$ is the temporal variance\wh{,} $\mu(\cdot)$ is the spatial mean, and $\mathbf{X}^j$ is a randomly selected sequence. All these data transformations maintain the action information.

\vspace{1mm}

$\bullet$ \textbf{Non-Actionlet Transformation $\mathcal{T}_{\text{non}}$}: 
To obtain stronger generalization, several extra data transformations are applied to the non-actionlet regions in addition to the above data transformation.

We apply an intra-instance data transformation \{Random Noise\} and an inter-instance data transformation \{Skeleton Mix\}. The random Noise has larger variance. Skeleton Mix is an element-wise data mixing method, including Mixup~\cite{zhang2017mixup}, CutMix~\cite{yun2019cutmix}, and ResizeMix~\cite{ren2022simple}. Because these transformations are performed on non-actionlet regions, they do not change the action semantics. Therefore, the transformed data are used as positive samples with the original data.

$\bullet$ \textbf{Actionlet-Dependent Combination}:
To merge the data transformations of the two regions, we utilize actionlets to combine.
It is formalized as:
\begin{equation}
    \label{equ:info}
    \begin{aligned}
    \mathbf{X}^i_{\text{trans}} = \mathbf{A}^i_{tv} \odot \mathbf{X}^i_{\text{act}} + (1 - \mathbf{A}^i_{tv}) \odot \mathbf{X}^i_{\text{non}},
    \end{aligned}
\end{equation}
where $\mathbf{X}^i_{\text{trans}}$ is the final transformed data, $\mathbf{X}^i_{\text{act}}$ and $\mathbf{X}^i_{\text{non}}$ are transformed with actionlet transformations $\mathcal{T}_{\text{act}}$. Then, $\mathbf{X}^i_{\text{non}}$ is performed non-actionlet transformations $\mathcal{T}_{\text{non}}$. $\mathbf{A}^i_{tv}$ represents the actionlet.
\vspace{1mm}

\noindent\textbf{Semantic-Aware Feature Pooling (SAFP).} 
\label{sec:safp}
To extract the motion information more accurately, we propose \wh{a} semantic-aware feature pooling method along the spatial-temporal dimension. This method focuses only on the feature representation of the actionlet region, thus reducing the interference of other static regions for motion feature extraction. It is formalized as:
\begin{equation}
    \label{equ:abfp}
    \begin{aligned}
     \text{SAFP}(\mathbf{h}^i_{ctv})= \sum_{t=1}^T \sum_{v=1}^V \mathbf{h}^i_{ctv} \left(\frac{\mathbf{A}^i_{tv}}{\sum_{t=1}^T \sum_{v=1}^V \mathbf{A}^i_{tv}}\right).
    \end{aligned}
\end{equation}
% where $\kappa$ is a hyper-parameter. 

This semantic-aware feature aggregation approach effectively extracts motion information and makes the features more distinguishable. We utilize this semantic-aware feature pooling operation in the offline stream to provide accurate anchor features.

% \vspace{1mm}

\begin{table*}[tb]
\small
\centering
\caption{Comparison of action recognition results with unsupervised learning approaches on NTU dataset.}
\begin{tabular}{l|c|c|c|c|c}
    \toprule
    % \hline
    Models& Stream &NTU 60 xview&NTU 60 xsub&NTU 120 xset&NTU 120 xsub\\
    \midrule
    AimCLR~\cite{guo2021contrastive} & joint & 79.7 & 74.3 & 63.4 & 63.4\\
    % CMD$^\dag$~\cite{mao2022cmd} & joint & 81.3 & 76.8 & 66.0 & 65.4 \\
    \textbf{ActCLR} & joint & \textbf{86.7} &\textbf{80.9}&\textbf{70.5} &\textbf{69.0}\\
    \midrule
    AimCLR~\cite{guo2021contrastive} & motion & 70.6 & 66.8 & 54.4 & 57.3\\
    % CMD$^\dag$~\cite{mao2022cmd} & motion &  80.4 & 75.8 & 64.8 & 64.2 \\
    \textbf{ActCLR} & motion & \textbf{84.4} &\textbf{78.6}&\textbf{67.8} &\textbf{68.3}\\
    \midrule
    AimCLR~\cite{guo2021contrastive} & bone & 77.0 & 73.2 & 63.4& 62.9\\
    % CMD$^\dag$~\cite{mao2022cmd} & bone & 80.7 & 75.2 & 65.0 & 64.7 \\
    \textbf{ActCLR} & bone & \textbf{85.0} &\textbf{80.1}&\textbf{68.2} &\textbf{67.8}\\
    \midrule
    3s-AimCLR~\cite{guo2021contrastive} & \quad joint+motion+bone \quad & 83.8 & 78.9 & 68.8 & 68.2\\
    % 3s-CMD$^\dag$~\cite{mao2022cmd} & \quad joint+motion+bone & 85.0 & 79.9 & 69.6 & 69.1\\
    \textbf{3s-ActCLR} & \quad joint+motion+bone \quad & \textbf{88.8} &\textbf{84.3}&\textbf{75.7} &\textbf{74.3}\\
    \bottomrule
\end{tabular}
\label{tab:unsupervised_ntu}
\end{table*}

\vspace{1mm}

\noindent\textbf{Training Overview.}
In this part, we conclude our framework of contrastive learning in detail: 
\begin{itemize}%[itemsep=2pt]
\setlength{\itemsep}{1.5pt} 
    \item[1)] Two encoders are pre-trained using MoCo v2~\cite{he2020momentum}, an online encoder $f_q(\cdot)$ and an offline encoder $f_k(\cdot)$.  \zjh{The online encoder is updated via back-propagation gradients, while the offline encoder is a momentum-updated version of the online encoder as described} in Sec.~\ref{sec:contra}.
    \item[2)] The offline network $f_k(\cdot)$ inputs the original data $\mathbf{X}^i$ and we employ the unsupervised actionlet selection module to generate actionlet regions $\mathbf{A}^i_{tv}$ in the offline stream in Sec.~\ref{sec:actionlet}.
    \item[3)] We perform data transformation $\mathcal{T}$ to obtain two different views $\mathbf{X}^i_q$ and $\mathbf{X}^i_k$. And we apply motion-adaptive transformation strategy (MATS) to enhance the diversity of $\mathbf{X}^i_q$ in Sec.~\ref{sec:mats}.
    \item[4)] For feature extraction, in online stream, $\mathbf{z}^i_q = (g_q \circ \text{GAP} \circ f_q \circ \text{MATS})(\mathbf{X}^i_q)$, where $g_q(\cdot)$ is an online projector and GAP is the global average pooling. To provide a stable and accurate anchor feature, we utilize the semantic-aware feature pooling (SAFP) method in Sec.~\ref{sec:safp} to generate offline features $\mathbf{z}^i_k = (g_k \circ \text{SAFP} \circ f_k)(\mathbf{X}^i_k)$, where $g_k(\cdot)$ is an offline projector. 
    % Two encoders are employed, an online encoder $f_q(\cdot)$ and an offline encoder $f_k(\cdot)$ to extract features: $\mathbf{z}^i_q = g_q(f_q(\mathbf{X}^i_q))$ and $\mathbf{z}^i_k = g_k(f_k(\mathbf{X}^i_k))$, where $g_q(\cdot)$ is an online projector and $g_k(\cdot)$ is an offline projector. The offline networks $f_k(\cdot)$ and $g_k(\cdot)$ are updated by the momentum of the online networks $f_q(\cdot)$ and $g_q(\cdot)$. by $\hat{f} \leftarrow \alpha \hat{f} + (1 - \alpha) f$, where $\alpha$ is a momentum coefficient. 
    \item[5)]  A memory bank $\mathbf{M} = \{\mathbf{m}^i\}^M_{i=1}$ is utilized to store offline features. The offline features extracted from the offline data in each batch are stored in the memory bank, and the bank is continuously updated using a first-in first-out strategy.
    \item[6)] Following recent works~\cite{mao2022cmd,zhang2022contrastive}, we exploit similarity mining to optimize:
    \begin{equation}
    \label{equ:info}
    \begin{aligned}
    &\mathcal{L}_{\text{KL}}(\mathbf{p}_q^i, \mathbf{p}_k^i) = -\mathbf{p}_k^i \log \mathbf{p}_q^i,\\
    &\mathbf{p}_q^i = \text{SoftMax}(\text{sim}(\mathbf{z}^i_q, \mathbf{M})/\tau_q),\\
    &\mathbf{p}_k^i = \text{SoftMax}(\text{sim}(\mathbf{z}^i_k, \mathbf{M})/\tau_k),
    \end{aligned}
    \end{equation}
    where $\text{sim}(\mathbf{z}^i_q, \mathbf{M}) = [\text{sim}(\mathbf{z}^i_q, \mathbf{m}^j)]^M_{j=1}$, which indicates the similarity distribution between feature $\mathbf{z}^i_q$ and other samples in $\mathbf{M}$. For the elements $\mathbf{p}_k^{ij}$ of $\mathbf{p}_k^i$ greater than the elements $\mathbf{p}_q^{ij}$ of $\mathbf{p}_q^i$, these corresponding features $\mathbf{m}^j$ in the memory bank are positive samples. This is because the network increases the similarity of the output with these features.
\end{itemize}

\begin{table}[tb]
    \small
    \centering
    \caption{Comparison of action recognition results with unsupervised learning approaches on NTU 60 dataset. $^\dag$ indicates that results reproduced on our settings of feature dimension size.}
    \begin{tabular}{l|c|c|c}
      \toprule
      % \hline
      Models&Architecture&xview&xsub\\
      \midrule
      \rowcolor{gray!10} \multicolumn{4}{l}{\textit{Single-stream:}}\\
      LongT GAN~\cite{zheng2018unsupervised} & GRU & 48.1 & 39.1\\
      MS$^2$L~\cite{lin2020ms2l} & GRU & - & 52.5\\
      AS-CAL~\cite{rao2021augmented} & LSTM & 64.8 & 58.5 \\
      P$\&$C~\cite{su2020predict} & GRU & 59.3 & 56.1 \\
      SeBiReNet~\cite{nie2020unsupervised} & SeBiReNet & 79.7 & -\\
      ISC~\cite{thoker2021skeleton}& GCN $\&$ GRU & 78.6 & 76.3 \\
      AimCLR~\cite{guo2021contrastive}& GCN & 79.7 & 74.3\\
      CMD$^\dag$~\cite{mao2022cmd} & GRU & 81.3 & 76.8 \\
      GL-Transformer~\cite{kim2022global} & Transformer & 83.8 & 76.3\\
      CPM~\cite{zhang2022contrastive} & GCN & 84.9 & 78.7\\
      \textbf{ActCLR} & GCN & \textbf{86.7} &\textbf{80.9}\\
      \midrule
      \rowcolor{gray!10} \multicolumn{4}{l}{\textit{Three-stream:}}\\
      3s-Colorization~\cite{yang2021skeleton} & DGCNN & 83.1 & 75.2 \\
      3s-CrosSCLR~\cite{li20213d} & GCN & 83.4 & 77.8\\
      3s-AimCLR~\cite{guo2021contrastive}& GCN & 83.8 & 78.9\\
      3s-CMD$^\dag$~\cite{mao2022cmd} & GRU & 85.0 & 79.9 \\
      3s-SkeleMixCLR~\cite{chen2022contrastive} & GCN & 87.1 & 82.7\\
      3s-CPM~\cite{zhang2022contrastive} & GCN & 87.0 & 83.2\\
      \textbf{3s-ActCLR} & GCN & \textbf{88.8} &\textbf{84.3}\\
      \bottomrule
    \end{tabular}
    \label{tab:unsupervised_ntu_60}
  \end{table}
  
  % \begin{table*}[tb]
  %   \caption{Comparison of action recognition results with unsupervised learning approaches on NTU 120 dataset.}
  %   \label{tab:unsupervised_ntu}
  %   \centering
  %   \begin{tabular}{l|c|c|c|c}
  %     \toprule
  %     % \hline
  %     Models&NTU 60 xview&NTU 60 xsub&NTU 120 xset&NTU 120 xsub\\
  %     \midrule
  %     MS$^2$L~\cite{lin2020ms2l} & - & 52.5 & - & -\\
  %     AS-CAL~\cite{rao2021augmented} & 64.8 & 58.5 & 49.2 & 48.6 \\
  %     P$\&$C~\cite{su2020predict} & 59.3 & 56.1 & 44.1 & 41.4 \\
  %     SeBiReNet~\cite{nie2020unsupervised} & 79.7 & - & - & -\\
  %     AimCLR~\cite{guo2021contrastive} & 79.7 & 74.3 & - & -\\
  %     3s-Colorization~\cite{yang2021skeleton} & 83.1 & 75.2 & - & -\\
  %     ISC~\cite{thoker2021skeleton} & 78.6 & 76.3 & \textbf{67.1} & \textbf{67.9}\\
  %     3s-CrosSCLR~\cite{li20213d} & 83.4 & \textbf{77.8} & 66.7 & \textbf{67.9}\\
  %     BCLR & \textbf{83.9} &\textbf{77.8}&66.8& 67.5\\
  %     \bottomrule
  % \end{tabular}
  % \end{table*}

\section{Experiment Results}

For evaluation, we conduct our experiments on the following two datasets: the NTU RGB+D dataset~\cite{shahroudy2016ntu,liu2019ntu} and the PKUMMD dataset~\cite{liu2020pku}. 
% Our goal is to obtain the feature encoder $f(\cdot)$ which can generate good feature representations for action recognition. 

\subsection{Datasets and Settings}
\noindent$\bullet$ \textbf{NTU RGB+D Dataset 60 (NTU 60)}~\cite{shahroudy2016ntu} is a large-scale dataset which contains 56,578 videos with 60 action labels and 25 joints for each body, including interactions with pairs and individual activities.

\vspace{1mm}

\noindent$\bullet$ \textbf{NTU RGB+D Dataset 120 (NTU 120)}~\cite{liu2019ntu} is an extension to NTU 60 and the largest dataset for action recognition, which contains 114,480 videos with 120 action labels. Actions are captured with 106 subjects with multiple settings using 32 different setups.

\vspace{1mm}

\noindent$\bullet$ \textbf{PKU Multi-Modality Dataset (PKUMMD)}~\cite{liu2020pku} covers a multi-modality 3D understanding of human actions. The actions are organized into 52 categories and include almost 20,000 instances. There are 25 joints in each sample. The PKUMMD is divided into part I and part II. Part II provides more challenging data, because the large view variation causes more skeleton noise.

To train the network, all the skeleton sequences are temporally down-sampled to 50 frames. The encoder $f(\cdot)$ is based on ST-GCN~\cite{yan2018spatial} with hidden channels of size 16, which is a quarter the size of the original model. The projection heads for contrastive learning and auxiliary tasks are all multilayer perceptrons, projecting features from 256 dimensions to 128 dimensions. $\tau_q$ is $0.1$ and $\tau_k$ is $0.04$. We employ a fully connected layer $\phi(\cdot)$ for evaluation.

\begin{table}[tb]
  \small
  \centering
  \caption{Comparison of action recognition results with unsupervised learning approaches on NTU 120 dataset. $^\dag$ indicates that results reproduced on our settings of feature dimension size.}
  \begin{tabular}{l|c|c|c}
    \toprule
    % \hline
    Models&Architecture&xset&xsub\\
    \midrule
    \rowcolor{gray!10} \multicolumn{4}{l}{\textit{Single-stream:}}\\
    AS-CAL~\cite{rao2021augmented} & LSTM & 49.2 & 48.6 \\
    AimCLR~\cite{guo2021contrastive} & GCN & 63.4 & 63.4\\
    CMD$^\dag$~\cite{mao2022cmd} & GRU & 66.0 & 65.4 \\
    GL-Transformer~\cite{kim2022global} & Transformer & 68.7 & 66.0\\
    CPM~\cite{zhang2022contrastive} & GCN & 69.6 & 68.7\\
    \textbf{ActCLR} & GCN & \textbf{70.5} &\textbf{69.0}\\
    \midrule
    \rowcolor{gray!10} \multicolumn{4}{l}{\textit{Three-stream:}}\\
    3s-CrosSCLR~\cite{li20213d} & GCN & 66.7 & 67.9\\
    3s-AimCLR~\cite{guo2021contrastive} & GCN & 68.8 & 68.2\\
    3s-CMD$^\dag$~\cite{mao2022cmd} & GRU & 69.6 & 69.1 \\
    3s-SkeleMixCLR~\cite{chen2022contrastive} & GCN & 70.7 & 70.5\\
    3s-CPM~\cite{zhang2022contrastive} & GCN & 74.0 & 73.0\\
    \textbf{3s-ActCLR} & GCN & \textbf{75.7} &\textbf{74.3}\\
    \bottomrule
  \end{tabular}
  \label{tab:unsupervised_ntu_120}
\end{table}

To optimize our network, Adam optimizer~\cite{newey1988adaptive} is applied, and we train the network on one NVIDIA TitanX GPU with a batch size of 128 for 300 epochs.

\begin{table*}[tb]
    \small
    \centering
    \caption{Comparison of action recognition results with supervised learning approaches on NTU dataset.}
    \begin{tabular}{l|c|c|c|c|c}
      \toprule
     Models&Params&NTU 60 xview&NTU 60 xsub&NTU 120 xset&NTU 120 xsub\\
      \midrule
      \rowcolor{gray!10} \multicolumn{6}{l}{\textit{Single-stream:}}\\
      % \textit{Single-stream:}\\
      ST-GCN~\cite{yan2018spatial} & 0.83M & 88.3 & 81.5 & 73.2 & 70.7 \\
      SkeletonCLR~\cite{li20213d} & 0.85M & 88.9 & 82.2 & 75.3 & 73.6\\
      AimCLR~\cite{guo2021contrastive} & 0.85M & 89.2 & 83.0 & 76.1 & 77.2\\
      CPM~\cite{zhang2022contrastive} & 0.84M & 91.1 & 84.8 & 78.9 & 78.4\\
      % \textbf{F4F (Parameter-Efficient)} & 0.23M & 90.1 & 84.7 & 79.3 & 76.6\\
      \textbf{ActCLR} & 0.84M & \textbf{91.2} &\textbf{85.8}&\textbf{80.9}& \textbf{79.4}\\
      \midrule
      \rowcolor{gray!10} \multicolumn{6}{l}{\textit{Three-stream:}}\\
      3s-ST-GCN~\cite{yan2018spatial} & 2.49M & 91.4 & 85.2 & 77.1 & 77.2 \\
      3s-CrosSCLR~\cite{li20213d}& 2.55M & 92.5 & 86.2 & 80.4 & 80.5 \\
      3s-AimCLR~\cite{guo2021contrastive}& 2.55M & 92.8 & 86.9 & 80.9 & 80.1\\
      3s-SkeleMixCLR~\cite{chen2022contrastive}& 2.55M & 93.9 & 87.8 & 81.2 & 81.6\\
      % \textbf{3s-F4F (Parameter-Efficient)} & 0.69M & 92.7 & 87.5 & 83.2 & 81.0\\
      \textbf{3s-ActCLR} & 2.52M & \textbf{93.9} &\textbf{88.2}&\textbf{84.6}& \textbf{82.1}\\
      \bottomrule
  \end{tabular}
    \label{tab:supervised_ntu}
  \end{table*}

\subsection{Evaluation and Comparison}
To make a comprehensive evaluation, we compare our method with other methods under variable settings.
\vspace{1mm}

\begin{table}[tb]
    \small
    \centering
    \caption{Comparison of the transfer learning performance on PKUMMD dataset with linear evaluation pretrained on NTU 60.}
    \begin{tabular}{l|c|c}
        \toprule
    Models&PKU I xview& PKU II xview\\
      \midrule
      3s-AimCLR~\cite{guo2021contrastive} & 85.3 & 42.4 \\
      \textbf{3s-ActCLR} & \textbf{91.6} &\textbf{44.5}\\
      \midrule
      \midrule
     Models& PKU I xsub& PKU II xsub\\
      \midrule
      LongT GAN~\cite{zheng2018unsupervised} & - & 44.8 \\
      MS$^2$L~\cite{lin2020ms2l} & - & 45.8\\
      ISC~\cite{thoker2021skeleton} & - & 51.1\\
      Hi-TRS~\cite{chen2022hierarchically} & - & 55.0\\
      3s-CrosSCLR~\cite{li20213d} & - & 51.3\\
      3s-AimCLR~\cite{guo2021contrastive} & 85.6 & 51.6 \\
      \textbf{3s-ActCLR} & \textbf{90.0} &\textbf{55.9}\\
      \bottomrule
  \end{tabular}
    \label{tab:trans_pku}
  \end{table}

\noindent\textbf{1) Linear Evaluation.}
In the linear evaluation mechanism, a linear classifier $\phi(\cdot)$ is applied to the fixed encoder $f(\cdot)$ to classify the extracted features. We adopt action recognition accuracy as a measurement. Note that this encoder $f(\cdot)$ is fixed in the linear evaluation protocol.

Compared with other methods in Tables~\ref{tab:unsupervised_ntu},~\ref{tab:unsupervised_ntu_60} and~\ref{tab:unsupervised_ntu_120}, our model shows superiority on these datasets. We find that the transformation that 3s-CrosSCLR~\cite{li20213d} and 3s-AimCLR~\cite{guo2021contrastive} design in the contrastive learning task is unified for different regions, which makes the data transformation interfere with the motion information. On the contrary, our method adopts MATS for semantic-aware motion-adaptive data transformation. Thus, the features extracted by our method maintain better action information which is more suitable for downstream tasks. 

\vspace{1mm}

\noindent\textbf{2) Supervised Finetuning.}
We first pretrain the encoder $f(\cdot)$ in the self-supervised learning setting, and then finetune the entire network. We train the encoder $f(\cdot)$ and classifier $\phi(\cdot)$ using complete training data.

Table~\ref{tab:supervised_ntu} displays the action recognition accuracy on the NTU datasets. This result confirms that our method extracts the information demanded by downstream tasks and can better benefit action recognition. 
In comparison with state-of-the-art supervised learning methods, our model achieves better performance.

\vspace{1mm}

% \noindent\textbf{\wh{3)} Semi-Supervised Approaches.}
% In semi-supervised learning, the training process utilizes both labeled data and unlabeled data. Generally, the encoder $f(\cdot)$ is pretrained with the contrastive learning task with full data, and then fine-tuned with the classifier $\phi(\cdot)$ with labeled data. To give a comprehensive and thorough evaluation, we conduct experiments under different settings, including $1\%$, $5\%$, $10\%$ and $20\%$ of labeled skeleton sequences. We also show the comparison results with the state-of-the-art methods. 

% In Table~\ref{tab:semi_supervised_pku} and~\ref{tab:semi_supervised_ntu}, we notice that with small subsets of the datasets, our method improves the accuracy considerably and performs better than the state-of-the-art methods. Especially with smaller training data, our method outperforms the state-of-the-art by large margins.

% \noindent\textbf{\wh{3)} Parameter-Efficient Finetuning.}
% We also first pretrain the encoder $f(\cdot)$, and after self-distillating the encoder $f(\cdot)$, employ the gradient-guided parameter-efficient finetuning.

% Table~\ref{tab:supervised_ntu} shows the action recognition accuracy on the NTU datasets. Our method achieves comparable performance to full model fine-tuning with fewer trainable parameters. Moreover, we can adjust the \wh{number} of parameters that need to be finetuned through the threshold, which makes the finetuning more controllable, as shown in Table~\ref{tab:effi}.

\vspace{1mm}

\noindent\textbf{3) Transfer Learning.} 
To explore the generalization ability, we evaluate the performance of transfer learning. In transfer learning, we exploit self-supervised task pretraining on the source data. Then we utilize the linear evaluation mechanism to evaluate on the target dataset. In linear evaluation, the encoder $f(\cdot)$ has fixed parameters without fine-tuning.

As shown in Table~\ref{tab:trans_pku}, our method achieves significant performance. Our method employs MATS to remove irrelevant information, and SAFP to retain information related to downstream tasks. This allows our encoder $f(\cdot)$ to obtain stronger generalization performance.

\vspace{1mm}

\noindent\textbf{4) Unsupervised Action Segmentation.} 
To explore the extraction of local features by our method, we used unsupervised action segmentation as an evaluation metric. We pre-train the encoder $f(\cdot)$ on the NTU 60 dataset. Then we utilize the linear evaluation mechanism to evaluate \wh{the results} on the PKUMMD dataset. In linear evaluation, the encoder $f(\cdot)$ has fixed parameters without fine-tuning.

As shown in Table~\ref{tab:seg_pkuII}, our method achieves significant performance. Because our method focuses on the main occurrence region of the action, it is possible to locate the actions out of the long sequence.

\begin{table}[tb]
    \small
    \begin{center}
    \caption{Comparison of the action segmentation performance on PKUMMD II xview dataset with linear evaluation pretrained on NTU 60 xview dataset.}
    \label{tab:seg_pkuII}
    \begin{tabular}{l|c|c|c|c|c}
    \toprule
    \multirow{2.5}*{Models}&\multirow{2.5}*{Stream}&\multicolumn{4}{c}{PKUMMD II xview}\\
    \cmidrule(lr){3-6}
    &&ACC & MACC & FWIoU & mIoU\\
    \midrule
    AimCLR~\cite{guo2021contrastive} & joint & 39.77 & 28.68 & 26.79 & 15.67\\
    \textbf{ActCLR} & joint & \textbf{51.29}  & \textbf{31.97} & \textbf{35.24} & \textbf{21.38}\\
    \midrule
    AimCLR~\cite{guo2021contrastive}& motion& 42.32 & 26.65 & 29.92 & 15.92 \\
    \textbf{ActCLR} & motion& \textbf{56.69} & \textbf{39.45} & \textbf{41.34} & \textbf{27.73}\\
    \midrule
    AimCLR~\cite{guo2021contrastive}& bone& 54.22 & 39.52 & 39.41 & 27.36\\
    \textbf{ActCLR} &bone& \textbf{59.09} & \textbf{41.14} & \textbf{41.54} & \textbf{28.89}\\
    \bottomrule
    \end{tabular}
    \end{center}
    \end{table}

\subsection{Ablation Study}
Next, we conduct ablation experiments to give a more detailed analysis of our proposed approach.

\vspace{1mm}

\noindent\textbf{1) Analysis of Motion-Adaptive Data Transformation.} 
Data transformation is very important for consistency learning. To explore the influence of motion-adaptive data transformations, we test the action recognition accuracy under different data transformations. 
As shown in Table~\ref{tab:data}, the motion-adaptive transformation can obtain better performance than full region (the whole skeleton data) in different noise settings. It is also observed that when the noise strength increases, our performance degradation is much smaller than that of full region. This indicates that the design is more robust to data transformation.
% the consistency of the feature space is further enhanced with actionlet-dependent data transformations. And under stronger data transformations, our method still maintains the motion semantic information. Accordingly, the performance of downstream tasks is improved. 

To explore the influence of different data transformations on the contrastive learning effect, we test the action recognition accuracy under different data transformation combinations. As shown in Table~\ref{tab:com}, the consistency of the feature space is further enhanced with more data transformations. Thus, the performance of the downstream task is improved.

% \begin{table}[tb]
% \small
% \begin{center}
% \caption{Analysis of actionlet-dependent data transformation on NTU 60 dataset with the joint stream.}
% \label{tab:data}
% \begin{tabular}{l|c|c|c|c}
% \toprule
% \multirow{2.5}*{Transformation}& \multicolumn{2}{c}{Region}&\multicolumn{2}{|c}{xview}\\
% \cmidrule(lr){2-5}
% & Non-Actionlet & Full Area & KNN & Linear \\
% \midrule
% \multirow{2}*{Noise 0.01} & \checkmark &  & 77.63 & 86.46 \\
% &  & \checkmark & 76.51 & 85.91 \\
% \midrule
% \multirow{2}*{Noise 0.05} & \checkmark &  & \textbf{78.04} & \textbf{86.79} \\
% &  & \checkmark & 75.28 & 84.20 \\
% \midrule
% \multirow{2}*{Noise 0.1} & \checkmark &  & 77.31 & 86.12\\
% &  & \checkmark & 74.19 & 83.69\\
% \midrule
% \multirow{2}*{Skeleton Mix} & \checkmark &  & \textbf{78.04} & \textbf{86.79}\\
% &  & \checkmark & 73.24 & 83.05\\
% \bottomrule
% \end{tabular}
% \end{center}
% \end{table}

\begin{table}[tb]
    \small
    \begin{center}
    \caption{Analysis of motion-adaptive data transformation on NTU 60 xview dataset with the joint stream.}
    \label{tab:data}
    \begin{tabular}{l|c|c|c}
    \toprule
    Transformation & Region & KNN & Linear\\
    \midrule
    \multirow{2}*{Noise 0.01} & Non-Actionlet & 77.63 & 86.46 \\
    &  Full Area & 76.51 & 85.91 \\
    \midrule
    \multirow{2}*{Noise 0.05} & Non-Actionlet  & \textbf{78.04} & \textbf{86.79} \\
    &  Full Area & 75.28 & 84.20 \\
    \midrule
    \multirow{2}*{Noise 0.1} & Non-Actionlet  & 77.31 & 86.12\\
    &  Full Area & 74.19 & 83.69\\
    \midrule
    \multirow{2}*{Skeleton Mix} & Non-Actionlet  & \textbf{78.04} & \textbf{86.79}\\
    &  Full Area & 73.24 & 83.05\\
    \bottomrule
    \end{tabular}
    \end{center}
    \end{table}
    
    \begin{table}[tb]
    \small
    \begin{center}
    \caption{Analysis of data transformation combinations on NTU 60 xview dataset with the joint stream. $\mathcal{T}$ is all the transformations. $\mathcal{T}_{\text{act}}$ is actionlet transformations. $\mathcal{T}_{\text{non}}$ is non-actionlet transformations. AdaIN refers to Skeleton AdaIN.}
    \label{tab:com}
    \begin{tabular}{l|c|c}
    \toprule
    % \multirow{2.5}*{Proportion}&\multicolumn{2}{|c}{xview}\\
    % \cmidrule(lr){2-3}
    Modules& KNN & Linear\\
    \midrule
    w/o $\mathcal{T}$ & 67.50 & 79.98\\
    w/o (AdaIN + $\mathcal{T}_{\text{non}}$) & 69.75 & 81.80\\
    w/o $\mathcal{T}_{\text{non}}$ & 73.63 & 83.27\\
    Full Version & \textbf{78.04} & \textbf{86.79}\\
    \bottomrule
    \end{tabular}
    \end{center}
    \end{table}

\vspace{1mm}

\noindent\textbf{2) Analysis of Semantic-Aware Feature Pooling.} 
To explore the semantic-aware feature pooling, we perform this pooling on different streams. Table~\ref{tab:effi} shows the results of accuracy of action recognition under different settings. We note that better performance is obtained with offline, as it makes offline to generate better positive sample features for contrastive learning. Using this module in online reduces the benefits exposed by the non-actionlet transformation.
% In offline network feature extraction, $\kappa$ controls the attention of the actionlet. A smaller $\kappa$ makes the feature pooling close to the global average pooling, while a larger $\kappa$ makes the features extracted only from the actionlet region.
% %
% Table~\ref{tab:effi} shows the results of different $\kappa$. 
% We notice that using only SAFP can get better performance than using GAP. And the combination of both is used to obtain more informative features.

\begin{table}[tb]
    \small
    \begin{center}
    \caption{Analysis of semantic-aware feature pooling on NTU 60 xview dataset with the joint stream.}
    \label{tab:effi}
    \begin{tabular}{l|c|c}
    \toprule
    % \multirow{2.5}*{Proportion}&\multicolumn{2}{|c}{xview}\\
    % \cmidrule(lr){2-3}
    Modules& KNN & Linear\\
    \midrule
    w/o SAFP & 76.38 & 85.69\\
    % 1.0 & 75.89 & 86.28\\
    offline w/ SAFP  & \textbf{78.04} & \textbf{86.79}\\
    online w/ SAFP & 76.02 & 85.25\\
    online + offline w/ SAFP & 76.71 & 85.92\\
    \bottomrule
    \end{tabular}
    \end{center}
    \vspace{-1em}
    \end{table}

\vspace{1mm}

% \noindent\textbf{3) Analysis of Unsupervised Actionlet Selection.} 
% We explored the effect of thresholding on actionlet selection for feature learning. As shown in Table~\ref{tab:threshold}, when the threshold is small, many joints are selected causing the actionlet to contain many static regions. When the threshold value is relatively large, the selected area is relatively small, making the motion information lost.
% \input{cvpr2023-author_kit-v1_1-1/latex/tab/thresholds.tex}

% \vspace{1mm}

\begin{figure}[tb]
    \centering
    \includegraphics[width=0.9\linewidth]{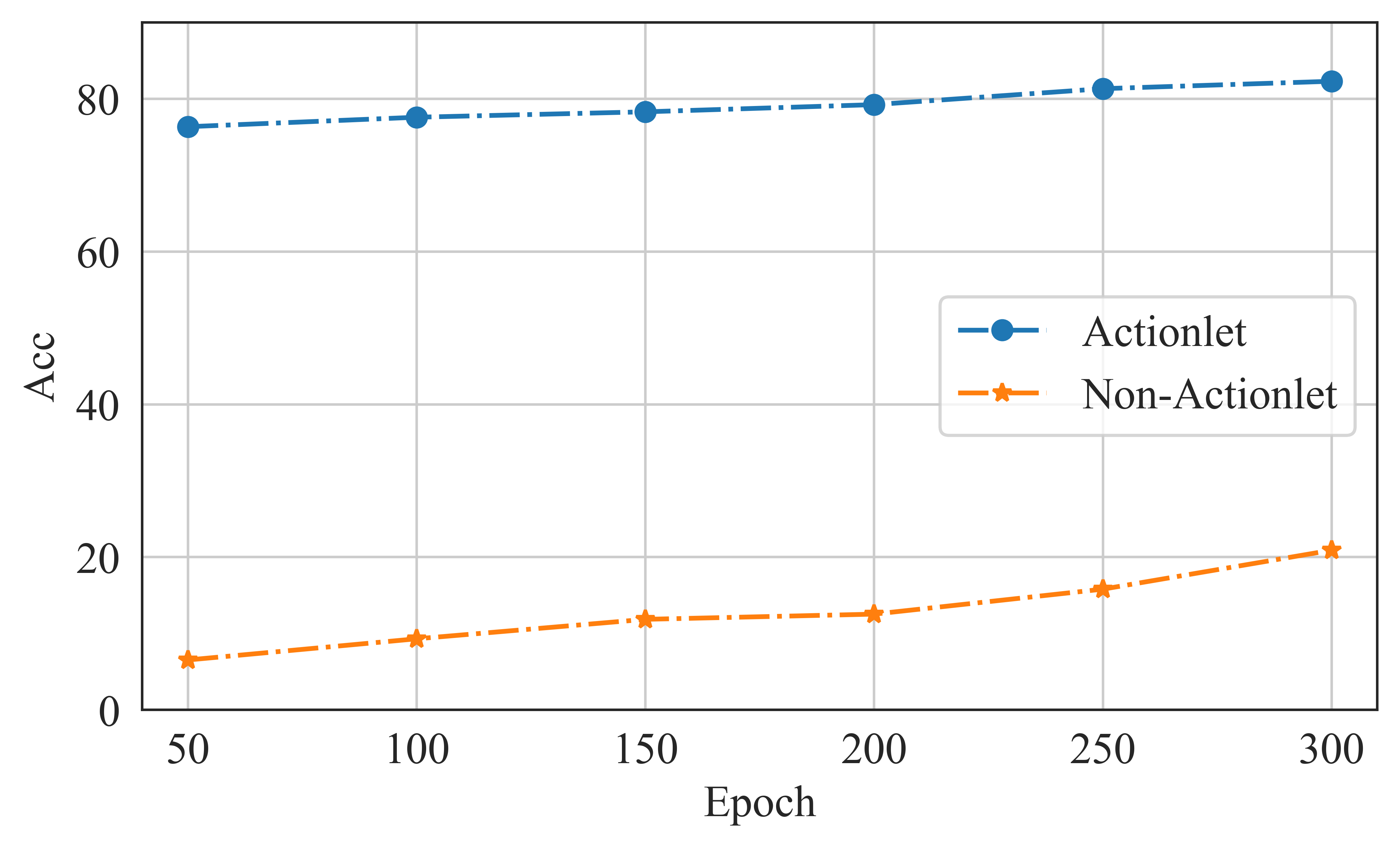}
    \caption{
    Action recognition accuracy of actionlet regions and non-actionlet regions. 
    }
    \label{fig:acc}
    \end{figure}

\noindent\textbf{3) Analysis of Actionlet and Non-Actionlet Semantic Decoupling.}   
In Fig.~\ref{fig:acc}, we show the performance of extracting only actionlet region information and non-actionlet region information for action recognition. The accuracy of the actionlet region for action recognition is comparable to the accuracy of the whole skeleton data. In contrast, the performance of the features of non-actionlet regions for action recognition is much lower. This shows that the actionlet area does contain the main motion information. 

\vspace{1mm}

\begin{figure}[tb]
    \includegraphics[width=\linewidth]{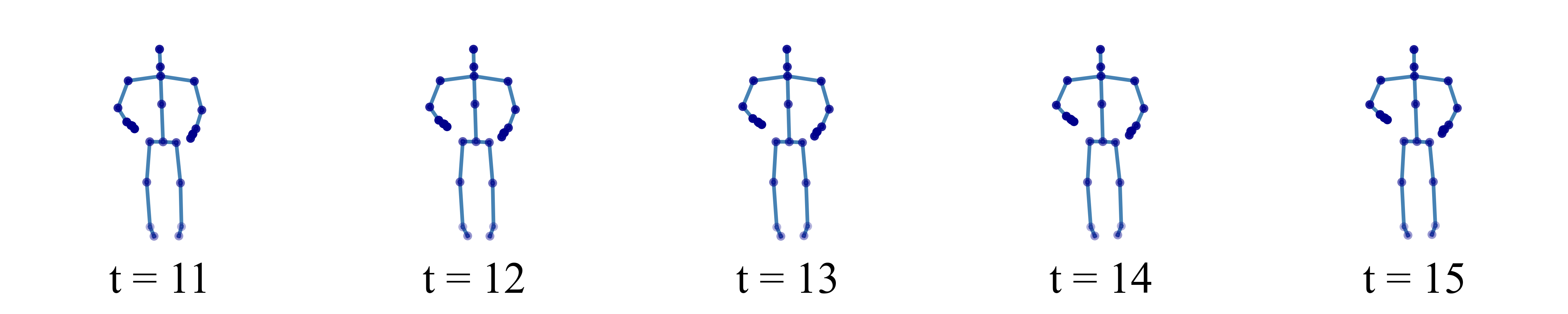}
    \caption{Visualization of the average motion. No obvious action takes place in the average motion sequence and can therefore be considered as a static anchor.}
    \label{fig:mean}
    \end{figure}

\begin{figure}[tb]
    \includegraphics[width=\linewidth]{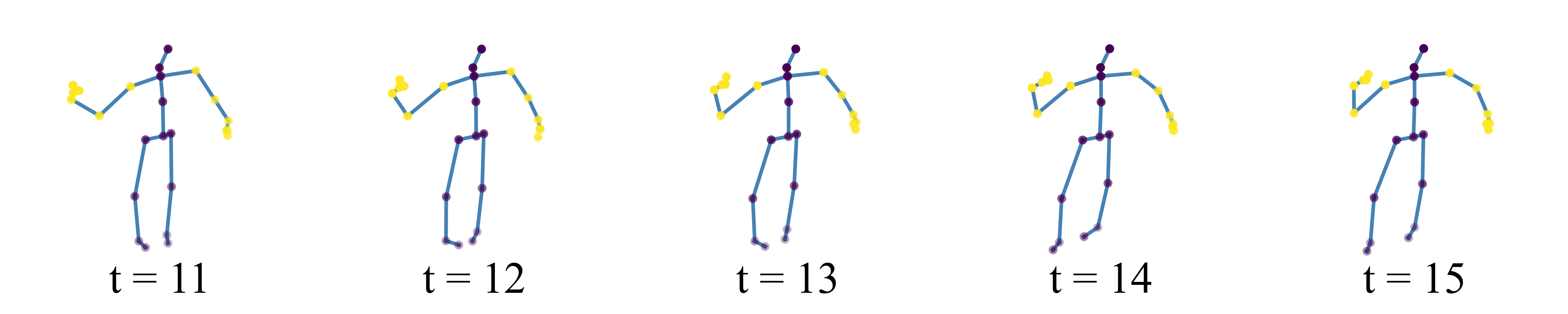}
    \caption{
    Visualization of the actionlet for a ``throw" sequence. The yellow joints are the actionlet. Note that hand movements are mainly selected, indicating that the actionlet is reasonable.
    }
    \label{fig:act}
    \end{figure}

\noindent\textbf{4) Visualization of Average Motion and Actionlet.} 
\wh{Fig}.~\ref{fig:mean} shows a visualization of the average motion and actionlet respectively. The average motion has no significant motion information and serves as a background. The actionlet, shown in Fig.~\ref{fig:act}, selects the joints where the motion mainly occurs. Our actionlet is spatio-temporal, because the joints with motion may change when the action is performed.

\section{Conclusions}
% \lorem{3}
In this work, we propose a novel actionlet-dependent contrastive learning method. Using actionlets, we design motion-adaptive data transformation and semantic-aware feature pooling to decouple action and non-action regions. These modules make the motion information of the sequence to be attended to while reducing the interference of static regions in feature extraction. In addition, the similarity mining loss further regularizes the feature space. Experimental results show that our method can achieve remarkable performance and verify the effectiveness of our designs.

{\small
\bibliographystyle{ieee_fullname}
\bibliography{ref}
}

\end{document}